# MAMMODINO: ANATOMICALLY AWARE SELF-SUPERVISION FOR MAMMOGRAPHIC IMAGES

*Sicheng Zhou, Lei Wu, Cao Xiao, Parminder Bhatia, Taha Kass-Hout*

GE HealthCare, Washington, US

## ABSTRACT

Self-supervised learning (SSL) has transformed vision encoder training in general domains but remains underutilized in medical imaging due to limited data and domain specific biases. We present MammoDINO, a novel SSL framework for mammography, pretrained on 1.4 million mammographic images. To capture clinically meaningful features, we introduce a breast tissue aware data augmentation sampler for both image-level and patch-level supervision and a cross-slice contrastive learning objective that leverages 3D digital breast tomosynthesis (DBT) structure into 2D pretraining. MammoDINO achieves state-of-the-art performance on multiple breast cancer screening tasks and generalizes well across five benchmark datasets. It offers a scalable, annotation-free foundation for multipurpose computer-aided diagnosis (CAD) tools for mammogram, helping reduce radiologists' workload and improve diagnostic efficiency in breast cancer screening.

## 1. INTRODUCTION

Breast cancer is the most prevalent cancer for women in the US [1], and the mammogram, an X-ray based imaging of the breast, remains the primary way of breast cancer screening and detection. Accurate interpretation of mammograms is essential for early diagnosis and effective treatment planning, but the task remains challenging even for experienced radiologists due to high inter-reader variability, subtle imaging characteristics of early-stage lesions, and the confounding effects of breast density. As a result, CAD tools have been developed to support radiologists by improving detection accuracy and reducing reading workload. Current CAD tools for mammogram were mainly developed based on traditional machine learning models using supervised learning schema, e.g., ResNet [2, 3, 4]. However, the lack of annotated mammogram data limits the generalizability and efficacy of these models.

To reduce the dependence on the scarce annotations, recent works in medical imaging have shifted toward foundation model pretraining. In this paradigm, large-scale vision encoders are first pretrained on large amount of unlabeled data and then adapted to specific downstream tasks by fine-tuning. Two dominant paradigms have emerged, the image-only SSL [5, 6, 7], and weakly supervised pretraining, such as text-guided contrastive learning [8, 9]. In medical domain, the pretrained vision encoders have been developed either on single or multiple modalities, such as X-Ray, computed tomography (CT) and magnetic resonance imaging (MRI). For instance, Pérez-García *et al.* [10] developed the RAD-DINO, a Vision Transformer (ViT) based encoder pretrained continually on large chest X-ray collections using image-only SSL. Despite the absence of text supervision, RAD-DINO achieves competitive or superior performance compared to strong image–text pre-training baselines on disease classification and segmentation tasks. Similarly, MedCoSS [11] proposes a sequential SSL framework that pretrains across heterogeneous modalities. Demonstrating that staged, continual SSL yields universal and robust visual representations. In contrast, text-guided pretraining aligns visual features with semantic meaning. For instance, BiomedCLIP [12] leverages millions of biomedical image-caption pairs in a CLIP style training scheme, achieving strong performance in zero/few-shot disease classification and visual question answering tasks. Mammo-CLIP [13] further leverages paired mammogram-report data to enhance the performance in breast density estimation, lesion classification and cancer screening tasks.

However, mammograms present several modality-specific challenges that are not fully addressed by generic SSL pipelines. The clinical signal of interest is often localized within breast tissue regions, while large background areas are typically non-informative. Furthermore, mammograms frequently involve 3D DBT, where context information is distributed across adjacent 2D slices. Conventional DINO styled SSL frameworks employ random cropping augmentations and contrastive learning on views of the same 2D image. Such designs are suboptimal for mammography. Random cropping may oversample irrelevant background. Contrastive learning constrained to individual 2D slices failing to capture the cross-slice structural coherence in 3D DBT data.

To address these limitations, we propose MammoDINO, a vision encoder for mammography built on the DINOv2 SSL framework, enhanced with two key innovations. First, we introduced breast tissue aware augmentation sampler, makes the augmented crops are constrained to breast tissue regions. It ensures that the model focuses on clinically meaningful areas during pretraining. Second, we designed the 3D DBT adjacent slice loss, a novel contrastive style loss that enforces consistency across adjacent 2D slices within the 3D DBT volume, effectively capturing anatomical continuity. These contributions align the self-supervised training signal with the physiological and geometric characteristics of mammography, resulting in more discriminative, robust and clinically meaningful vision representations for downstream breast cancer screening tasks, such as cancer detection, lesion detection, BI-RADS score prediction, and breast density classification.

## 2. METHODOLOGY

### 2.1. MammoDINO pretraining overview

DINOv2 [6] based SSL was adopted to pretrain the MammoDINO encoders. **Figure 1** shows the overview of the study. We developed the breast tissue aware DINO loss and iBOT loss (DINO-M and iBOT-M). Also, we designed and integrated the 3D DBT adjacent

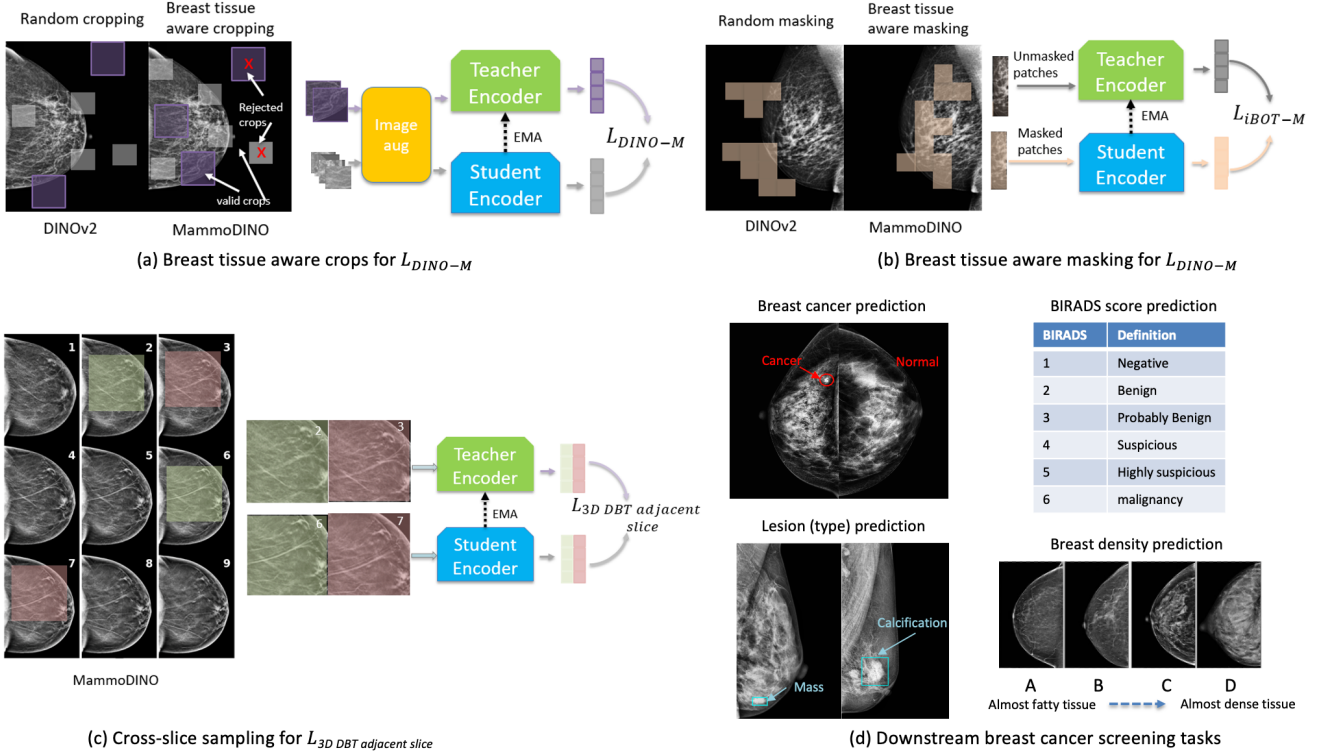

Fig. 1. Overview of proposed SSL training schema for MammoDINO. (a) Breast tissue aware sampler applied to obtain informative crops for image level DINO-M loss calculation. (b) Breast tissue aware masking for the patch-level iBOT-M loss calculation. (c) Adjacent slice pairs (slices 2, 6 and slices 3, 7) are sampled from the 3D DBT volume. The slice pairs are processed with conservative breast tissue aware transforms for further DINO-adj loss calculation. (d) Detailed illustrations of the downstream breast cancer screening tasks for model evaluations. The tasks include binary and multi-class classifications.

slice loss (DINO-adj) into the overall loss function:

$$L_{total} = \lambda_1 \sum L_{DINO-M} + \lambda_2 \sum L_{iBOT-M} + \lambda_3 \sum L_{DINO-adj} + \lambda_4 \sum L_{KoLeo}$$

### 2.2. Breast tissue aware DINO loss

Regular DINO loss takes random global and local crops as inputs for the student and teacher models, for mammograms, this may over sample the background area that has no information. We designed the breast tissue aware crop sampler that replaces the random crops, enforce a minimum breast tissue fraction in every crops. **Figure 1(a)** shows its comparison with DINOv2 random sampler and overview of the DINO-M loss. Concretely, given a mammogram image $x \in R^{H \times W}$, a binary tissue mask $M$ is created by 1) min-max normalization of the image $\hat{x} = (\tilde{x} - min\tilde{x})/(max\tilde{x} - min\tilde{x})$; 2) Let $\tau$ be a fixed percentile, define the intensity threshold $\theta = percentile(\hat{x}, \tau)$ and get the raw mask $M_0 = \mathbb{1}[\hat{x} > \theta]$; 3) denoise with morphological closing and opening using a 9×9 kernel to obtain the breast tissue mask $M \in \{0,1\}^{H \times W}$. For any cropped window $C$, it's breast tissue coverage is $r(C) = |C|^{-1} \sum M_{(i,j)}$ where (i,j) $\in C$. The cropped window is placed around a random pixel within the breast tissue, and the window is valid only if the $r(C)$ is larger than the minimum breast coverage ratio $\rho$. By enforcing $r(C) > \rho$, the supervisory signal consistently targets meaningful breast tissue area rather than background, yielding representations that transfer better to mammography detection and classification. Let $T(\cdot)$ be the designed breast tissue aware view sampler, $X_s = T(x)$ and $X_t = T(x)$ are the breast tissue aware student and teacher view sets, respectively. The student/teacher CLS token probabilities for a cropped view $u$ are $p_s(u) = softmax\left(\frac{h_s(f_s(u))}{\tau_s}\right)$ and $p_t(u) = softmax\left(\frac{h_t(f_t(u))-c}{\tau_t}\right)$ where $f_s/f_t$ are student/teacher encoders and $h_s/h_t$ are student/teacher MLP heads; $\tau_s/\tau_t$ are temperatures and $c$ is the centering vector (only for teacher) for normalization. The breast tissue aware DINO loss (DINO-M) is defined as:

$$L_{DINO-M} = -\frac{1}{|X_s||X_t|} \sum_{u \in X_s} \sum_{v \in X_t} \left(\sum_{k=1}^{K} p_t(v)_k \log p_s(u)_k\right)$$

### 2.3. Breast tissue aware iBOT loss

Similarly, to make the patch-level objective focus on clinically informative regions, we replaced uniform rectangular masking with a breast tissue aware sampler that operates on the masked image modeling (MIM) grid ($H_m \times W_m$ image patches) induced by the crop size and patch size. **Figure 1(b)** shows its comparison with DINOv2 masking. Given a mammogram image $x$, and a target of $m$ masked patches, we first build a binary tissue mask M as described in previous section. The sampler then adds mask pieces (block of image patches) until m tokens are masked. For each piece, it samples a target area $a_t \in [m_{min}, m_{max}]$ (in patches) and scores every feasible top-left grid coordinates $(r, c)$ by its breast tissue coverage ratio $r_{r,c}$, coordinates not meet the minimum breast tissue coverage ratio $r_{r,c} < \rho$ are discarded. For remaining coordinates, assign a sampling weight $\pi_{r,c} = w_t r_{r,c} + \varepsilon$ to make windows with more breast coverage tissue are more likely to be chosen while keep the randomness. The chosen window $\hat{C}$ is accepted if it contributes new coverage $0 < |\hat{C}| - o \leq \min(m_{max}, m - c)$ beyond current mask $Z$ (overlap $o$) and its pixels are set to 1 in $Z$. If no valid coordinates are found to fulfil $m$ masked patches, the breast coverage ratio threshold is relaxed linearly. The resulting mask $Z \in \{0,1\}^{H_m \times W_m}$ defines the index set $\Omega = \{(i,j): Z_{ij} = 1\}$ on which the iBOT loss is computed, i.e., a cross-entropy between teacher patch codes and student predictions restricted to $(i,j) \in \Omega$. The breast tissue aware iBOT loss is defined as:

$$L_{iBOT-M} = -\frac{1}{|\Omega|} \sum_{(i,j) \in \Omega(x)} \left( \sum_{k=1}^{K} p_t(x)_k^{i,j} \log p_s(x)_k^{i,j} \right)$$

### 2.4. 3D DBT adjacent slice loss

3D DBT volumes comprise closely spaced 2D slices in which anatomical structures change smoothly along the depth axis. Standard image-only SSL (global/local crops and MIM) does not explicitly exploit this cross-slice structural coherence. We therefore designed the 3D DBT adjacent slice loss that aligns the model's CLS token predictions between a pair of nearby slices from the same DBT volume. Specifically, given a DBT volume $V$ with total $K$ slices, a slice index $k$ and an offset $d \in \{1, \dots, D_{max}\}$, a pair of slices $V_k$ and $V_{k \pm d}$ are sampled. **Figure 1(c)** shows examples of two slice pairs. To preserve correspondence, both slices are processed with a conservative, breast tissue aware sampler $T(\cdot)$ described in section 2.2. Two views are formed: $x_k = T(V_k)$ and $x_{k'} = T(V_{k \pm d})$. The student encoder $f_s$ and MLP head $h_s$ takes $x_k$ to obtain student CLS token logits $C_s$, similarly, teacher encoder $f_t$ and MLP head $h_t$ takes $x_{k'}$ to obtain teacher CLS token logits $C_t$. The teacher logits are further centered and temperature sharpened to teacher probabilities, $p_t(C_t) = softmax((C_t - c)/\tau_t)$, student probabilities are calculated in similar way without centering. $p_s(C_s) = softmax(C_t/\tau_s)$. The adjacent slice loss is a cross-entropy that pulls the student's distribution on slice $V_k$ toward the teacher's distribution on slice $V_{k \pm d}$:

$$L_{DINO-adj} = -\sum_{k=1}^{K} p_t(C_t)_k \log p_s(C_s)_k$$

## 3. EXPERIMENTS AND RESULTS

### 3.1. MammoDINO pretraining datasets

We collected both 2D reconstructed mammograms and 3D DBT from the clinical sites across United States and Europe. The 2D mammograms include standard mediolateral oblique and bilateral craniocaudal views. The 3D DBT were further processed into 2D slices. In summary, the pretraining dataset contains 1,400,323 2D images, consists of 42,386 reconstructed 2D mammograms and 1,357,937 2D slices. All pretraining mammogram images were pre-processed for three steps: 1) transformed to 8-bit from 16-bit; 2) min-max normalization; 3) contrast limited adaptive histogram equalization [14].

### 3.2. MammoDINO model pretraining

We pretrained the MammoDINO variants based on the ViT-base architecture with patch size of 14. We set the input image size to (518, 518) and positional embeddings are interpolated to this resolution. We trained the models for 300,000 steps with AdamW, weight decay scheduled from 0.04 to 0.20 and batch size of 128. All experiments were performed on a single AWS EC2 instance with 8 NVIDIA L40S GPUs.

### 3.3. Benchmark datasets and evaluation results

We evaluated the breast cancer screening downstream tasks on 5 major public mammogram benchmark datasets, i.e., RSNA [15], VinDr-Mammo [16], DDSM [17], CMMD [18] and CDD-CESM [19]. The evaluation tasks include cancer detection (binary), lesion detection (binary), lesion type prediction (multiclass), BIRADS score prediction (multiclass) and breast density prediction (multiclass). Area Under the Curve (AUC) and F1 score are used for the evaluation metrics for binary and multi-class classification tasks, respectively. We compared our MammoDINO model with 1) supervised CNN based models, i.e., ResNet50 [20] and ConvNeXt models [21]; 2) generic SSL vision encoder DINOv2 [6]; 3) radiology-tailored SSL vision encoder RadDINO [10]; 4) weakly supervised (text-guided) pretrained medical vision encoders, i.e., BiomedCLIP [12] and MammoCLIP [13]. All models use the same 3-layer MLP classification head, appended to the backbone and fine-tuned on the benchmark datasets. The evaluation results for the five datasets are shown in **Table 1-5**.

**Table 1**: Model evaluations on VinDr dataset.

| | Cancer detection | Lesion detection | Lesion type prediction | BIRADS score prediction | Breast density prediction |
|---|---|---|---|---|---|
| ResNet50 | 0.554 | 0.676 | 0.315 | 0.518 | 0.793 |
| ConvNext | 0.853 | 0.674 | 0.333 | 0.534 | 0.821 |
| DINOv2 | 0.837 | 0.690 | 0.331 | 0.533 | 0.811 |
| RadDINO | 0.717 | 0.624 | 0.337 | 0.522 | 0.824 |
| BiomedCLIP | 0.764 | 0.550 | 0.302 | 0.518 | 0.788 |
| MammoCLIP | 0.809 | 0.668 | 0.313 | 0.518 | 0.802 |
| MammoDINO | **0.918** | **0.712** | **0.365** | **0.566** | **0.835** |

**Table 2**: Model evaluations on DDSM dataset.

| | Cancer detection | Lesion detection | BIRADS score prediction | Breast density prediction |
|---|---|---|---|---|
| ResNet50 | 0.716 | 0.919 | 0.664 | 0.619 |
| ConvNext | 0.708 | 0.910 | 0.667 | 0.660 |
| DINOv2 | 0.728 | 0.902 | 0.666 | 0.654 |
| RadDINO | 0.650 | 0.850 | 0.591 | 0.617 |
| BiomedCLIP | 0.681 | 0.779 | 0.561 | 0.396 |
| MammoCLIP | 0.694 | 0.878 | 0.635 | 0.650 |
| MammoDINO | **0.776** | **0.932** | **0.686** | **0.674** |

**Table 3**: Model evaluations on CDD-CSEM dataset.

| | Cancer detection | Lesion detection | BIRADS score prediction |
|---|---|---|---|
| ResNet50 | 0.452 | 0.538 | 0.165 |
| ConvNext | 0.512 | 0.534 | 0.426 |
| DINOv2 | 0.745 | 0.652 | 0.433 |
| RadDINO | 0.730 | 0.549 | 0.424 |
| BiomedCLIP | 0.597 | 0.513 | 0.165 |
| MammoCLIP | 0.624 | 0.809 | 0.373 |
| MammoDINO | **0.761** | **0.824** | **0.535** |

**Table 4**: Model evaluations on CMMD dataset.

|            | Cancer detection | Lesion detection | Lesion type prediction | BIRADS score prediction |
|------------|------------------|------------------|------------------------|-------------------------|
| ResNet50   | 0.659            | 0.493            | 0.271                  | 0.633                   |
| ConvNext   | 0.708            | 0.708            | 0.397                  | 0.695                   |
| DINOv2     | 0.711            | 0.688            | 0.410                  | 0.704                   |
| RadDINO    | 0.680            | 0.402            | 0.402                  | 0.680                   |
| BiomedCLIP | 0.637            | 0.271            | 0.271                  | 0.605                   |
| MammoCLIP  | 0.686            | 0.686            | 0.391                  | 0.688                   |
| MammoDINO  | **0.761**        | **0.737**        | **0.450**              | **0.762**               |

**Table 5**: Model evaluations on RSNA dataset.

|            | Cancer detection | BIRADS score prediction | Breast density prediction |
|------------|------------------|-------------------------|---------------------------|
| ResNet50   | 0.496            | 0.600                   | 0.719                     |
| ConvNext   | **0.649**        | **0.631**               | 0.733                     |
| DINOv2     | 0.620            | 0.617                   | 0.734                     |
| RadDINO    | 0.616            | 0.608                   | 0.740                     |
| BiomedCLIP | 0.458            | 0.586                   | 0.691                     |
| MammoCLIP  | 0.403            | 0.626                   | 0.748                     |
| MammoDINO  | 0.631            | 0.621                   | **0.759**                 |

Tables 1-5 shows that MammoDINO achieves the best performance for all tasks in five benchmark datasets except the cancer detection and BIRADS score prediction for RSNA. The **Figure 2** presents a comprehensive comparison of the seven models across all evaluation tasks and datasets. For each model, the value on a task axis is the mean metric of all five benchmark datasets. MammoDINO encloses the largest area, reflecting consistent gains on all downstream breast cancer screening tasks.

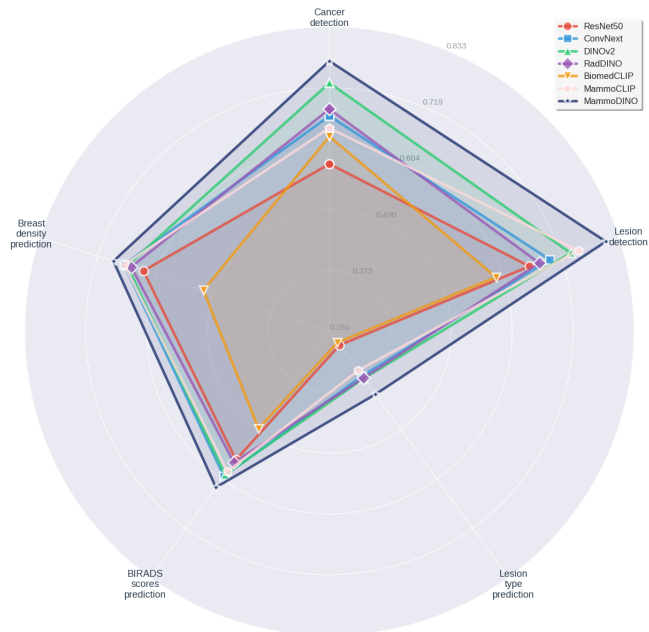

Fig. 2. Comprehensive comparison across models and tasks using a radar plot. The value on a task axis is the mean metric of all five benchmark datasets.

**Figure 3.** shows visualizations of two mammograms with identified suspicious lesion regions and the corresponding feature heatmaps. The visualizations highlight the features extracted by the MammoDINO are highly aligned with the breast tissue areas with clinical significance.

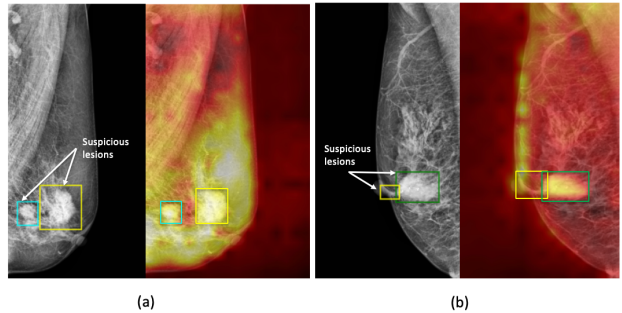

Fig. 3. Visualization of mammograms with annotated suspicious lesions and corresponding feature heatmaps

### 3.4. Ablation study

We quantify the contribution of the proposed components using four variants built on the ViT-B backbone: (a) DINOv2 baseline, (b) DINOv2 + breast tissue aware DINO/iBOT (+DINO-M & iBOT-M), (c) DINOv2 + 3D DBT adjacent slice loss, and (d) the full MammoDINO. We evaluated the variants using the five benchmark datasets and chose the most common three downstream tasks, i.e., cancer detection, lesion detection and BIRADS score prediction. The average performance of the same task across five datasets is calculated and showed in **Table 6**. The results indicate performance improvement of all three tasks when adding the DINO-M, iBOT-M and 3D DBT adjacent slice loss modules during pretraining, which further validate the effectiveness of proposed pretraining schema.

**Table 6**: Ablation studies of proposed DINO-M, iBOT-M and 3D DBT adjacent slice loss.

|                        | Cancer detection | Lesion detection | BIRADS score prediction |
|------------------------|------------------|------------------|-------------------------|
| DINOv2                 | 0.728            | 0.738            | 0.595                   |
| +DINO-M & iBOT-M       | 0.751 (**+0.23**)| 0.779 (**+0.41**)| 0.616 (**+0.21**)       |
| +3D DBT adjacent slice | 0.746 (**+0.18**)| 0.768 (**+0.30**)| 0.612 (**+0.17**)       |
| MammoDINO              | 0.769            | 0.801            | 0.634                   |

### 5. CONCLUSIONS

In this paper, we present MammoDINO, a vision encoder tailored for mammography, addressing key limitations of existing SSL frameworks in this domain. By introducing a breast tissue aware augmentation strategy and a novel 3D DBT adjacent slice loss, MammoDINO aligns the pretraining process with the anatomical and structural characteristics of mammography. These innovations enable the model to learn clinically relevant and robust visual representations without relying on manual annotations. Overall, our approach achieves the state-of-the-art performance for breast cancer screening tasks across benchmark datasets and provides a scalable, generalizable backbone for developing more effective CAD systems in mammographic imaging. Although the training schema is instantiated for mammograms, it is modality-agnostic and can benefit the vision encoder pretraining for other medical imaging modalities like CT and MRI.